\documentclass[letterpaper, 10 pt, conference]{ieeeconf}  % Comment this line out if you need a4paper

\IEEEoverridecommandlockouts                              % This command is only needed if 
\usepackage{graphics} % for pdf, bitmapped graphics files
\usepackage{epsfig} % for postscript graphics files
\usepackage{amsmath} % assumes amsmath package installed
\usepackage{amssymb}  % assumes amsmath package installed
\usepackage{tabularx}
\usepackage{booktabs}
\usepackage{cleveref}
\newcommand{\ie}{i.e.\ }
\newcommand{\eg}{e.g.\ }

\title{\LARGE \bf
Mean-Flow based One-Step Vision-Language-Action
}

\author{Yang Chen, Xiaoguang Ma, Bin Zhao}

\begin{document}

\maketitle
\thispagestyle{empty}
\pagestyle{empty}

%%%%%%%%%%%%%%%%%%%%%%%%%%%%%%%%%%%%%%%%%%%%%%%%%%%%%%%%%%%%%%%%%%%%%%%%%%%%%%%%
\begin{abstract}

Recent advances in FlowMatching-based Vision-Language-Action (VLA) frameworks have demonstrated remarkable advantages in generating high-frequency action chunks, particularly for highly dexterous robotic manipulation tasks. Despite these notable achievements, their practical applications are constrained by prolonged generation latency, which stems from inherent iterative sampling requirements and architectural limitations. To address this critical bottleneck, we propose a Mean-Flow based One-Step VLA approach. Specifically, we resolve the noise-induced issues in the action generation process, thereby eliminating the consistency constraints inherent to conventional Flow-Matching methods. This significantly enhances generation efficiency and enables one-step action generation. Real-world robotic experiments show that the generation speed of the proposed Mean-Flow based One-Step VLA is 8.7 times and 83.9 times faster than that of SmolVLA and Diffusion Policy, respectively. These results elucidate its great potential as a high-efficiency backbone for VLA-based robotic manipulation.

\end{abstract}

%%%%%%%%%%%%%%%%%%%%%%%%%%%%%%%%%%%%%%%%%%%%%%%%%%%%%%%%%%%%%%%%%%%%%%%%%%%%%%%%
\section{INTRODUCTION}

\begin{figure*}[t]
  \centering
  \includegraphics[width=\linewidth,height=0.32\textheight,keepaspectratio]{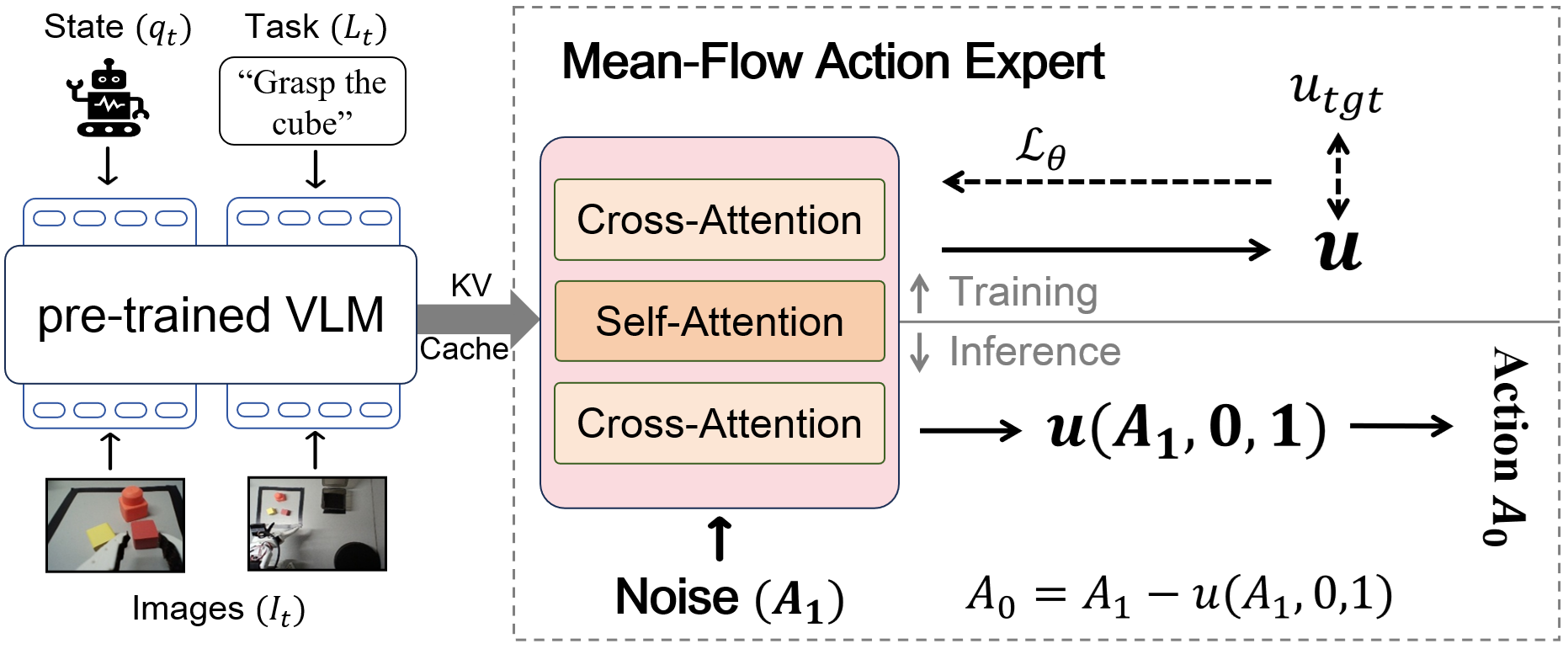}
  \caption{Overview of the Mean-Flow based One-Step VLA framework. The pretrained VLM processes multimodal inputs. During training, the Mean-Flow action expert approximates the mean denoising vector field conditioned on VLM features and predicts the mean vector field to obtain the Action $A_0$ during inference.}
  \label{fig:model}
\end{figure*}

Vision–Language–Action (VLA) models integrate semantic grounding, spatial reasoning, and sequential action generation into a unified framework, allowing robots to perform generalizable and adaptive behaviors across diverse scenarios \cite{sapkota2025vision}.
According to the adopted policy-learning paradigm, VLA models can be roughly classified into three categories: Transformer-based, Diffusion-based, and FlowMatching-based VLA \cite{kawaharazuka2025vision}.

The first VLA model, RT-1 \cite{brohan2022rt} by Google, is Transformer-based and is trained on 130k+ robotic demonstrations, enabling end-to-end mapping of visual inputs and achieving 90\%+ success rates on household tasks. RT-2 \cite{zitkovich2023rt} fuses pretrained vision-language models (VLMs), integrating web-scale world knowledge to generalize to unseen objects and abstract instructions without retraining, boosting cross-task performance by 30\%+ over RT-1 \cite{shao2025large}. Despite their successes in long-sequence modeling and unified multimodal representation, Transformer-based VLA models have critical disadvantages in discretization artifacts, motion discontinuities, and limited action diversity.

Instead of framing action generation as a token-based sequence prediction task and outputting discrete action tokens as in Transformer-based VLA, Diffusion-based VLA formulates robot visuomotor control as a denoising diffusion process and learns action-score gradients via stochastic Langevin dynamics wherein continuous action spaces are modeled directly through gradual noise reduction \cite{ho2020denoising}. The open-source Octo \cite{team2024octo} establishes the foundation for generalist robotic policies, demonstrating that Diffusion-based action generation can significantly improve zero-shot performance, \eg, 33\% over RT-1-X \cite{vuong2023open}.

FlowMatching-based VLA extends Diffusion-based approaches by directly generating continuous trajectories through flow-based dynamics. Unlike diffusion models that rely on iterative denoising, FlowMatching learns deterministic transport mappings between data and noise distributions, offering near-instant sampling for low latency, outperforming diffusion’s iterative denoising \cite{liu2022flow}. The FlowMatching-based action expert in $\pi_0$ \cite{black2024pi_0} reformulates the generative process as a continuous flow from the noise distribution to the data distribution, parameterized by a time-dependent vector field that satisfies a deterministic ordinary differential equation (ODE). The denoising process in FlowMatching is implemented through Euler integration over small discrete time intervals. When the number of inference steps is reduced, each Euler step covers a larger time span, amplifying the bias and causing the generated samples to drift further toward the dataset mean \cite{frans2024one}.
Consequently, FlowMatching faces an inherent trade-off between real-time efficiency and action generation accuracy, \ie, reducing the inference steps improves latency but compromises trajectory fidelity, whereas maintaining quality necessitates more integration steps and thus higher inference cost.

Unlike FlowMatching, which learns instantaneous vector fields and relies on multi-step numerical integration to reconstruct trajectories, MeanFlow learns a mean denoising vector field that directly models the overall mapping from the noisy endpoint to the data endpoint, thereby mitigating the accumulation of numerical errors and avoiding the need for multi-step iterative denoising inference \cite {geng2025mean}. Benefiting from this design, it requires no pre-training, distillation, or additional consistency heuristics.
In this paper, we introduce Mean-Flow based One-Step VLA to achieve high generation quality, low inference cost, and strong training stability simultaneously.
In fact, it can generate high-quality continuous actions in a single or few inference steps, significantly reducing both inference steps and latency while maintaining a simple training pipeline.

The contributions of this paper are :
\begin{itemize}
  \item We introduce Mean-Flow based One-Step VLA, a novel Vision-Language-Action framework that integrates the MeanFlow generative approach for robotic manipulation.
  \item We achieve significant efficiency gains by replacing the conventional FlowMatching mechanism with a mean denoising vector field predictor. The generation speed is 8.7 times faster over the state-of-the-art methods.
  \item We validate the framework through real-world robotic experiments, demonstrating that Mean-Flow based One-Step VLA achieves robust performance under both single-step and multi-step generation modes.
\end{itemize}

\section{RELATED WORK}

\subsection{Diffusion-based Visual-Language-Action Models}

Integrating Diffusion-based policies into VLA systems shifts robotic action generation from deterministic regression to multi-modal generative policies \cite{chi2025diffusion}.
Octo \cite{team2024octo} is the first Vision–Language–Action (VLA) model to adopt a Diffusion-based policies for continuous action control. It employs a Transformer backbone coupled with a Diffusion head that decodes action chunks through a denoising diffusion process to yield substantially improved performance and temporal coherence in action decoding.
While Diffusion-based policies generate smooth and high-frequency control signals, they lack the reasoning mechanisms necessary for solving complex tasks. Conventional autoregressive VLA models exhibit strong reasoning capabilities but often fail to produce precise and stable actions. Diffusion-VLA (DiVLA) \cite{wen2024diffusion} unifies these complementary strengths by coupling a large-scale pre-trained vision–language Model (VLM) for task decomposition and explanation with a diffusion head that performs robust and continuous action generation. In a zero-shot bin-picking benchmark involving 102 previously unseen objects, DiVLA achieves 63.7\% task success, demonstrating strong generalization to novel visual scenes.

\subsection{FlowMatching-based Visual-Language-Action Models}

FlowMatching is a family of generative models that construct flow paths and transport prior distribution to data distribution \cite{albergo2022building}. 
$\pi_0$ model adopts the FlowMatching framework, where training learns the vector fields. During inference, the model integrates this learned vector field over time to compute the trajectory from noise to the target state.
Compared with autoregressive prediction, FlowMatching is more efficient for continuous action generation, as it enables parallel prediction of entire action chunks in a single forward pass, thereby reducing inference latency \cite{zhang2025pure}.
By reducing model scale and trainable parameters, SmolVLA \cite{shukor2025smolvla} achieves an improvement in computational efficiency while maintaining performance comparable to much larger VLAs. It is designed to be trained on a single GPU and deployed on consumer-grade hardware, including CPUs, with alternative self-attention and cross-attention layers.

Nonetheless, FlowMatching-based VLA models still suffer from quality degradation when reducing the inference steps. This inherent limitation represents a key bottleneck for efficient and real-time generative control in currently available flow-based VLA systems.

\subsection{Generation Efficiency Improvement}

In diffusion models, the Number of Function Evaluations (NFE) denotes how many times the numerical solver calls the neural network. This directly determines the inference speed and computational cost. 
Existing efforts to reduce NFE primarily rely on ODE-based formulations \cite{lu2022dpm}, as deterministic dynamics are more tractable for numerical solvers. Alternative strategies such as distillation \cite{zhou2024score, salimans2022progressive, luo2023diff} or consistency models \cite{song2023consistency, lu2024simplifying, yang2024consistency} further reduce NFE but require additional training or complex scheduling. Recently, FlowMatching \cite{lipman2022flow} has emerged as a continuous-time alternative that constructs probability paths from noise distribution to data distribution without score estimation, achieving faster and smoother generation.
However, FlowMatching can only reduce the NFE to around 10, since further decreasing NFE leads to severe degradation in generation quality due to accumulated discretization errors. This limitation arises because the learned instantaneous vector field must be closely integrated along highly nonlinear probability paths, making coarse numerical updates inaccurate. 

Building upon the deterministic ODE framework, MeanFlow method \cite{geng2025mean} addresses the intrinsic limitation of FlowMatching by learning to approximate the mean denoising vector field over a time interval rather than the instantaneous one. MeanFlow predicts the average flow direction between two temporal states, capturing the overall trend of the probability path. The mean denoising vector field can be regarded as the integral of the instantaneous field, providing a mathematically rigorous formulation grounded in continuous-time dynamics.
By focusing on the coarse-grained evolution of the trajectory, MeanFlow narrows the gap between one-step diffusion/flow models and their multi-step predecessors. As a result, it substantially improves inference efficiency and simplifying the training pipeline without additional distillation or consistency regularization.

\section{METHODOLOGY}

This section describes how the MeanFlow principle is integrated into the Vision–Language–Action framework to improve both efficiency and stability of generation. We first derive the mathematical formulation of MeanFlow, then detail its instantiation within the VLA pipeline, and finally describe the proposed action-generation strategy. \cref{fig:model} illustrates the overall model architecture, and depicts the logic of Mean-Flow action expert.

\subsection{Introducing MeanFlow into VLA} 
The MeanFlow method is a flow-based generative approach designed to enhance the stability and efficiency of generative models through denoising techniques. The core idea is to introduce a novel ground-truth field, representing the global mean denoising vector field $u(z,r,t)$ over an interval $[r,t]$ \cite{sheng2025mp1}.
The mean vector field characterizes the average transport from noisy distribution to data distribution over a given interval, by aggregating the instantaneous denoising vector fields across time. This differs fundamentally from FlowMatching, which models the instantaneous denoising fields itself. This naturally serves as a fundamental principle guiding network training. A corresponding loss function is introduced to encourage the network to satisfy this intrinsic relationship between the mean and instantaneous vector fields. No additional consistency constraints are required, since the ground-truth target field is explicitly defined, the optimal solution is theoretically independent of the specific network architecture. This in practice results in more stable and robust training.

We define the mean vector field between two time steps $r$ and $t$ as the average transformation across the interval, obtained by aggregating the instantaneous denoising vector fields over time. Formally, the mean vector field $u$ is given by
\begin{equation}
    u(z_t, r, t) \triangleq \frac{1}{t - r} \int_r^t v(z_r, \tau) \, d\tau
  \label{eq:1}
\end{equation}
$z_t$ denotes the intermediate latent variable that continuously interpolates between the data sample $x$ and a noise sample $e$. It is defined as $z_t = a_tx + b_te$, with $e \sim \mathcal{N}(0, I)$.
The mean vector field is denoted by \(u\), and the instantaneous vector fields by \(v\). \(u\) is the result of a functional of \(v\). 

\begin{equation}
    u = \mathcal{F}[v] \triangleq \frac{1}{t - r} \int_r^t v \, d\tau
  \label{eq:2}
\end{equation}
Directly adopting the mean vector field defined in Eq.~\eqref{eq:2} as the ground-truth label for network training is impractical, therefore, we reformulate Eq.~\eqref{eq:2} as
\begin{equation}
    (t-r)u(z_t, r, t) = \int_r^t v(z_r, \tau) \, d\tau
  \label{eq:3}
\end{equation}
Differentiating both sides of the equation with respect to \( t \) yields an identical relation.
\begin{equation}
    u(z_t, r, t) + (t-r) \frac{d}{dt} u(z_t, r, t) = v(z_t, t)
  \label{eq:4}
\end{equation}
Therefore,
\begin{equation}
u(z_t, r, t) = v(z_t, t) - (t - r) \frac{d}{dt} u(z_t, r, t)
  \label{eq:5}
\end{equation}
This equation is called the ``MeanFlow Identity'', which describes the relation between $v$ and $u$.
The last term in Eq.~\eqref{eq:5} can be rewritten as
\begin{equation}
\frac{d}{dt} u(z_t, r, t) = \frac{dz_t}{dt} \partial_z u + \frac{dr}{dt} \partial_r u + \frac{dt}{dt} \partial_t u
  \label{eq:6}
\end{equation}
Let
\begin{equation}
\frac{dz_t}{dt} = v(z_t, t), \quad \frac{dr}{dt} = 0, \quad \frac{dt}{dt} = 1
  \label{eq:7}
\end{equation}
Eq.~\eqref{eq:7} can be rewritten as
\begin{equation}
\frac{d}{dt} u(z_t, r, t) = v(z_t, t) \partial_z u + \partial_t u
  \label{eq:8}
\end{equation}
Eq.~\eqref{eq:8} can be rewritten in matrix-multiplication form as
\begin{equation}
\frac{d}{dt} u(z_t, r, t) = [\partial_z u, \partial_r u, \partial_t u] [v(z_t, t), 0, 1]^T
  \label{eq:9}
\end{equation}
The first term is the derivative of the output with respect to each input variable and is called the Jacobian matrix. The second item is the instantaneous vector field and constants. Their product is referred to as the Jacobian-vector product (JVP), which is readily available in Jax and Torch.

We parameterize a network \(u_\theta\) to satisfy the MeanFlow Identity, enabling it to predict the mean vector field that transports samples from the noise distribution to the data distribution.
\begin{equation}
\mathcal{L}(\theta) = \mathbb{E} \left\| u_{\theta}(z_t, r, t) - sg(u_{\text{tgt}}) \right\|_2^2
  \label{eq:10}
\end{equation}
In Eq.~\eqref{eq:10}, \( u_{\text{tgt}} \) denotes the target mean vector field obtained from the ground-truth trajectory
\begin{equation}
u_{\text{tgt}} = v(z_t, t) - (t - r) (v(z_t, t) \partial_z u_{\theta} + \partial_t u_{\theta})
  \label{eq:11}
\end{equation}
and the stop-gradient operator $sg(·)$ is applied to prevent backpropagation through the JVP term, thereby avoiding higher-order gradient computations. Without this operation, gradients will propagate through the derivative of $u_\theta$, and do double backpropagation. This is computationally expensive and prone to instability\cite{geng2025mean}.

The MeanFlow approach can be viewed as FlowMatching with a modified target. When $r$ is always set to $t$ it reduces to standard FlowMatching. Whenever $r \neq t$, the model acquires the ability to generate in a single step. In practice, the network is required to fit both the instantaneous vector field and the mean vector field during training.

We aim to model the data distribution $p(A | O_t)$, where $A = \left[ \alpha_t, \alpha_{t+1}, \ldots, \alpha_{t+H} \right]$ represents the future action sequence and $O$ is an observation features. Each observation consists of multiple RGB images, a language command, and the robot’s proprioceptive state, denoted as $O_t = \left[ I_t^1, \ldots, I_t^n, L_t, q_t \right]$, where $I_t^i$ is the i-th image (two or three per robot), $L_t$ is the sequence of language tokens, and $q_t$ is the joint-angle vector. The images $I$ and state $q$ are encoded by respective encoders and then linearly projected into the same embedding dimension as the language tokens. The resulting embeddings are annotated with $(t, t - r)$, representing time and interval information.

For every action \(\alpha_t\) within the action chunk $A$, we obtain a corresponding action token conditioned on the observation, provided by the action expert. During training, these tokens are supervised by the MeanFlow loss, which replaces the conventional FlowMatching loss employed in prior work
\begin{equation}
\mathcal{L}(\theta) = \mathbb{E} \left\| u_{\theta}(A_t| O_t) - sg(u_{\text{tgt}}(A | O_t)) \right\|_2^2
  \label{eq:12}
\end{equation}
In the Mean-Flow based One-Step VLA model, we employ this framework to generate action trajectories $A_t$. Time steps $r$ and $t$ are randomly sampled to train the network $u_\theta(A_t, r, t)$ to predict the mean vector field, which then transports noise $A_1$ into the action $A_0$ trajectory
\begin{equation}
\mathcal{L}(\theta) = \mathbb{E} \left\| u_{\theta}(A_t, r, t) - sg\left( u_{\text{tgt}} \right) \right\|_2^2
  \label{eq:13}
\end{equation}

\subsection{Action Generation Strategy}
In the Mean-Flow based One-Step VLA model, the design of the action-generation policy is critical. To guarantee that the robot can execute tasks in real time within complex environments, we adopt a continuous-action generation strategy driven by MeanFlow. Vision, language, and proprioceptive states are first encoded and fused by the VLM backbone to produce a unified multimodal representation, as shown in \cref{fig:model}. The action expert then performs one-step denoising conditioned on this representation, directly generating the entire continuous action sequence for immediate execution. This yields trajectories that are both continuous and dynamically consistent. A major advantage of this design is its capacity to adapt the generative policy online as task conditions change, allowing fast reaction to novel visual inputs and enhancing real-time control performance.

At inference, actions are generated via the learned vector field from $\tau=0$ to $\tau=1$. Starting from random noise $A_1 \sim \mathcal{N}(0, I)$, single-step generation yields
\begin{equation}
A_0 = A_1 - u_\theta(A_1 | O_1)
  \label{eq:14}
\end{equation}
If multi-step generation is adopted, we obtain
\begin{equation}
A_{\gamma-\delta} = A_\gamma - \delta u_\theta(A_\gamma | O_\gamma )
  \label{eq:15}
\end{equation}
where \( \delta \) is the step size.

Sampling with the MeanFlow model simply replaces time integration with the mean vector field
\begin{equation}
A_r = A_t - (t - r) u_\theta(A_t, r, t)
  \label{eq:16}
\end{equation}
In the single-step sampling case, we simply have
\begin{equation}
A_0 = A_1 - u_\theta(A_1, 0, 1)
  \label{eq:17}
\end{equation}

\section{EXPERIMENTS}

\begin{figure*}[t]
  \centering
  \includegraphics[width=\linewidth,height=0.32\textheight,keepaspectratio]{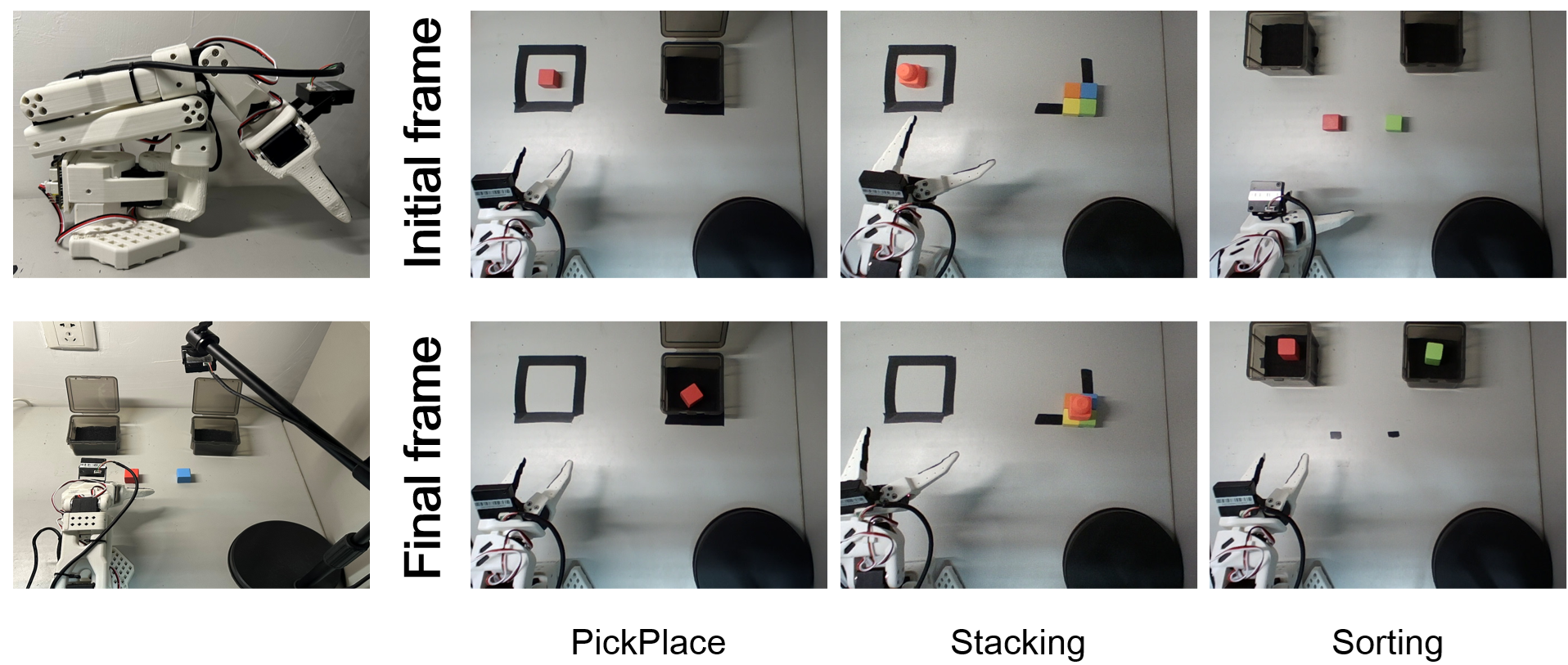}
  \caption{Illustrations of Real-world robotic arm and three real manipulation tasks.}
  \label{fig:expsetting}
\end{figure*}

Our experiments consist of two parts: a parameter sensitivity study and an ablation study.
Parameter analysis reveals how key hyperparameters affect performance and stability, to assess key design choices behind the optimal Mean-Flow based One-Step VLA model.
For the ablation evaluation, we compare our approach against SmolVLA and Diffusion Policy \cite{chi2025diffusion} across three real manipulation tasks.

Model Architecture :
We use SmolVLM-2 \cite{marafioti2025smolvlm}, an efficient vision–language model designed for multi-image and video inputs. It employs SigLIP \cite{zhai2023sigmoid} as the visual encoder and SmolLM-2 \cite{allal2025smollm2} as the language decoder, and fuses modalities via cross-attention between visual and textual tokens.

Mean-Flow action expert :
The Mean-Flow action expert $u_\theta$ is built upon a Transformer \cite{vaswani2017attention} architecture that predicts the mean vector field from VLM features, thereby generating continuous action trajectories $A = \left[ \alpha_t, \alpha_{t+1}, \ldots, \alpha_{t+H} \right]$.

Environment settings : 
All experiments are conducted on a home-made SO-101 robotic arm with six degrees-of-freedom (DoF) and a gripper. Two RGB cameras are used to cover the workspace, \ie, one mounted on the robot wrist for an ego-centric view and one fixed above the bench for a top-down third-person view, as shown in \cref{fig:expsetting}. To keep dataset and inference conditions identical, the physical scene is keep unchanged. Action generation speed is measured after a 20-minute idle period. During this interval, CPU and GPU temperatures are kept below 50 °C. Ambient temperature is maintained at approximately 15 °C with active ventilation and a stabilized power supply.

Tasks setting :
We collect three task-specific datasets using the SO-101 robot arm, each containing 100 demonstrations. For every demonstration, the initial objects positions and orientations are randomized to ensure diverse scene configurations.
We evaluate our method on three real manipulation tasks \ie, pick-place, stacking, and sorting, as shown in \cref{fig:expsetting}
\begin{itemize}
  \item For the pick-place task, the robot is instructed to \textit{``grasp the red block and put it in the box''}. The cube’s initial position and orientation are randomized in the 10 cm $\times$ 10 cm region. The score assigns 0.5 for a successful grasp and 0.5 for a correct placement.
  \item For stacking, the robot is instructed to \textit{``put the orange toy on the coloured blocks''}. The toy’s initial position and orientation are randomized in the same 10 cm $\times$ 10 cm area. The score again allocates 0.5 for a successful grasp and 0.5 for a successful stacking.
  \item The sorting task involves a longer horizon. The robot is instructed to \textit{``put the red cube in the left box and the green cube in the right box''}. Boxes locations remain fixed, and both cubes start with randomized orientations. The score allocates 0.25 for each successful grasp and 0.25 for each correct cube–box match. 
\end{itemize}

Training settings : 
Expert demonstrations are collected by manually teleoperating the robotic arm using a Joy-Con controller. Each trajectory lasts for at least 20 seconds, ensuring sufficient temporal coverage of task execution. The final dataset contains 300 trajectories.
At each timestep, the model receives a single-frame observation consisting of multi-view RGB images, a language instruction, and the robot’s proprioceptive state. The action space is 7-dimensional, including 6-DoF arm joint commands and 1 gripper control signal. Given the current observation, the model predicts a chunk of future actions in a multi-step manner, enabling temporally coherent motion generation.
The VLM backbone is frozen, and only the action expert is trained from scratch.
For the parameter sensitivity experiments, each model is trained for 60,000 steps.
For the ablation experiments, we adopt a multi-task training setup with 200,000 steps, while single-task training uses 60,000 steps. 
Training is performed on NVIDIA RTX 4090. Data collection and on-line inference are executed on a laptop equipped with an Intel Core i5-10200H and NVIDIA GTX 1650 GPU.

\subsection{Parameter Sensitivity}

We use a simple pick-place task to examine how different hyperparameters affect the performance of the Mean-Flow based One-Step VLA. For each hyperparameter configuration, we run 10 independent rounds, and in every round the model attempts 20 consecutive trials.
In practice, the optimal hyperparameter configuration for the Mean-Flow based One-Step VLA is unknown, and direct training under the 1-NFE setting proves highly unstable, often failing to converge. Therefore, we first conduct experiments under a more easily fitted 5-NFE setup to analyze the effects of the Flow Ratio and Loss Metric on model performance.

\begin{figure}[htbp]
  \centering
  \includegraphics[width=\columnwidth]{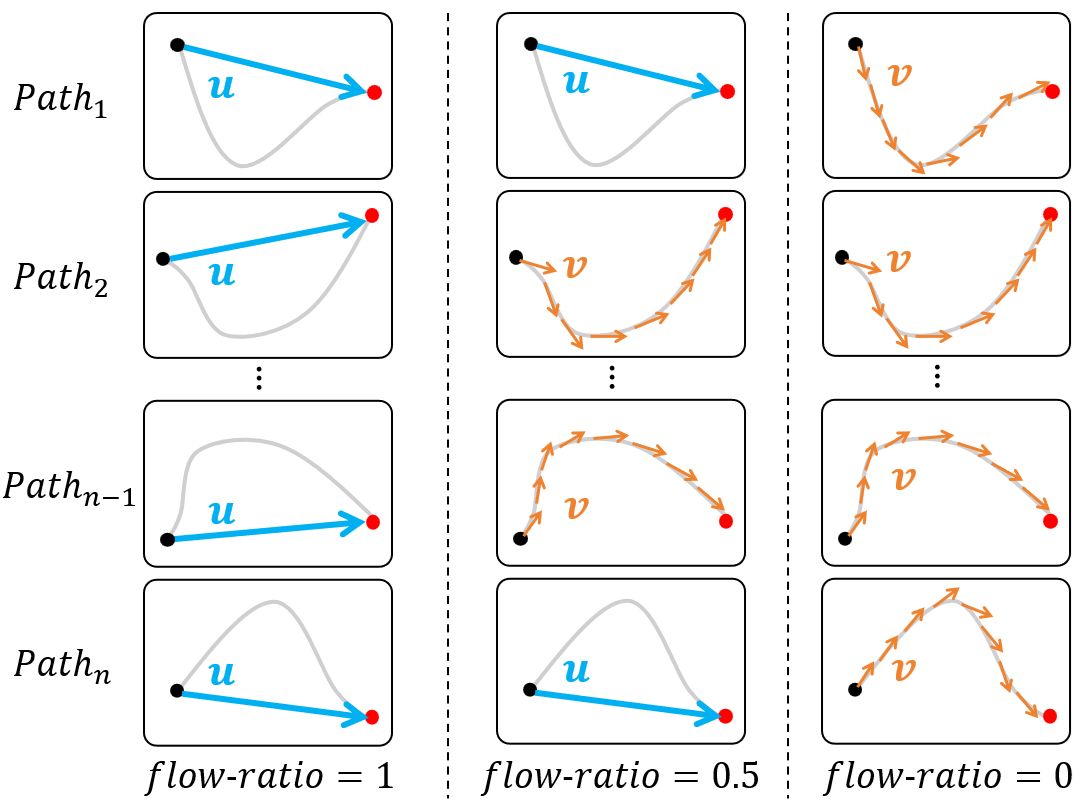}
  \caption{MeanFlow under various $flow\text{-}ratio$.}
  \label{fig:flowratio}
\end{figure}

\subsubsection{Flow Ratio}

During training, time steps \(r\) and \(t\) are randomly sampled (\(0\le r < t\le 1\)) so that the network learns the mean vector field between any two instants. To reinforce temporal continuity and vector field modeling, we introduce a flow-ratio hyperparameter, defined as
\begin{equation}
flow\text{-}ratio = 1 - \frac{N_{r=t}}{N_{total}}
  \label{eq:18}
\end{equation}
$N_{r=t}$ denotes the number of samples forced to be identical, \ie, training objective degenerates into predicting the instantaneous denoising vector field, as in standard FlowMatching, and $N_{total}$ is the total number of sampled time pairs within a batch. As shown in \cref{fig:flowratio}.

In fact, FlowMatching-based VLA suffers from severe degradation in action-generation quality when the number of inference steps is low. This limitation prevents further reduction in NFE. Therefore, in our experiments, we evaluate three flow ratios 0.2, 0.5, and 1.0 to examine how the flow ratio influences model performance under the NFE = 5 setting.

In \cref{tab:ratio} we find that the success rate is optimal at a ratio of 0.2. This setting balances learning between instantaneous and mean vector fields. Training only on the mean field loses instantaneous information and causes local trajectory errors, whereas training only on the instantaneous field struggles to guarantee overall correctness under one-step or few-step sampling.
In another word, the instantaneous vector field characterizes fine-grained and short-term variations in the action trajectory, while the mean vector field captures coarse-grained and long-term motion trends. Together, these complementary representations form a unified multi-scale description of visuomotor evolution, enabling both precise local control and stable global behavior during action generation.
The MeanFlow identity contains a time-derivative correction term $-(t-r)du/dt$, where $du/dt$ is computed via JVP (see \cref{eq:5}). Although the JVP is numerically exact for the current network parameters, it can amplify gradient variance for long-interval samples and introduce optimization instability during training. Introducing a certain proportion of $r = t$ samples makes the correction term vanish for those samples, thus providing local anchors for the target vector field and helping the network maintain a stable reference amid high-variance mean-flow samples. This improves training robustness and convergence stability. Furthermore, under few-step generation the model must perform large cross-interval transformations and preserve stepwise accuracy. Moderate $r = t$ sampling supplies supervision for local accuracy and complements learning of the cross-interval mean field, jointly reducing numerical integration error and improving task success rate.

\begin{table}[htbp]
\centering
\caption{Success rates (\%) under various flow ratios (NFE = 5).}
\label{tab:ratio}
\begin{tabular}{@{}cc@{}}
\toprule
Ratio & Success Rates\\
\midrule
0.2 & 84.5\\
0.5 & 80.5\\
1.0 & 4.5\\
\bottomrule
\end{tabular}
\end{table}

\subsubsection{Loss Metric}

Although $L_2$ loss is used in the previous \cref{eq:12}, training with this objective yield poor results and, in many runs, fail to converge. We therefore replace the $L_2$ loss with an Adaptive Loss, after which optimization stabilizes and final accuracy improves remarkably.
\begin{equation}
\mathcal{L} = \frac{1}{B} \sum_{i=1}^{B} \left[ \frac{1}{(\|\ \Delta_i \|^2 + c)^{1-\gamma}} \| \ \Delta_i \|^2 \right]
  \label{eq:19}
\end{equation}
In the equation, $ \|\ \Delta_i \|^2 $ represents the sum of squared errors for the i-th sample, \( \gamma \) is a hyperparameter controlling the degree of adaptivity, and \( c \) is a small constant to prevent division from being zero.

In \cref{tab:gamma} we evaluate the effect of the adaptive-loss hyperparameter $\gamma$ using three settings 0.3, 0.5, and 1.0. As $\gamma$ approaches 1.0, the loss function exerts the weakest down-weighting on outliers, and in the limit reverts to a plain $L_2$ loss. The results show that $L_2$ loss $(\gamma = 1)$ severely degrades performance, while the model achieves the highest success rate when $\gamma = 0.5$.

This trend can be explained by the multimodal nature of VLA data and the variance characteristics of the MeanFlow target. For the same visual–language condition, VLA data often correspond to multiple plausible action modes. A pure $L_2$ loss tends to regress toward the mean direction across these modes, leading to blurred or weakened predictions. Conversely, an excessively small $\gamma$ overly suppresses large-error samples, weakening supervision on the dominant modes. Moreover, since the MeanFlow target includes a time-derivative correction term proportional to $(t - r)$, samples from longer intervals exhibit higher target variance. The adaptive scaling controlled by $\gamma$ mitigates these effects by downweighting unstable and high-variance samples, thereby stabilizing optimization.

Setting $\gamma = 0.5$ provides a balanced trade-off between robustness and precision. It suppresses the gradient contribution of outliers while maintaining effective supervision on inlier samples, enabling the optimizer to focus on the main distribution. As a result, the training becomes more stable, converges faster, and achieves the highest overall success rates.

\begin{table}[htbp]
\centering
\caption{Success rates (\%) under various Gamma value with $flow\text{-}ratio$=0.2 (NFE = 5).}
\label{tab:gamma}
\begin{tabular}{@{}cc@{}}
\toprule
Gamma & Success Rates\\
\midrule
0.3 & 79.5 \\
0.5 & 86 \\
1.0 & 9.5 \\
\bottomrule
\end{tabular}
\end{table}

\subsubsection{Number of Function Evaluations(NFE)}

In \cref{tab:nfe} we evaluate the influence of NFE on model performance. With NFE = 1, the model can generate actions in a single step. However, the resulting trajectories are less accurate. As NFE increases from 1 to 5, the task success rate rises significantly. Raising NFE to 5 yields noticeably more accurate and stable motions, while a further increase to 10 provides only marginal improvement, indicating that a moderate number of evaluations offers the best balance between quality and computational cost.

This trend can be explained from both theoretical and empirical perspectives. When NFE = 1, the model must predict a single global mean vector field that transports the entire distribution from noise to data in one shot. Perfectly learning this global mapping is ideal, but in practice the dataset aggregates diverse demonstrations across multiple tasks, and the predicted horizon is long (\ie, the model predict up to 50 future action steps in a single forward pass), making such one-step estimation inherently difficult. Increasing NFE effectively partitions the time interval [0, 1] into smaller segments, forcing the model to learn local mean vector field over shorter intervals. Each local target is simpler and less nonlinear.  Therefore, it is necessary to strike a balance between model fitting error and solver integration error.

The improvement from NFE = 1 to NFE = 5 mainly arises from the intrinsic characteristics of the underlying vector field. The true marginal flow is highly nonlinear and induces curved trajectories. Multiple evaluation steps allow the model to capture simpler local conditional flows and mode-specific dynamics, resulting in smoother trajectories and higher success rates. In fact, NFE = 5 provides an effective trade-off, preserving MeanFlow’s efficiency while substantially enhancing motion fidelity, whereas further subdivision yields negligible gains due to limits in model capacity, dataset diversity, and numerical precision.

\begin{table}[htbp]
\centering
\caption{Success rates (\%) under various NFE values with $flow\text{-}ratio$=0.2, Gamma=0.5.}
\label{tab:nfe}
\begin{tabular}{@{}c *{5}{c} @{}}
\toprule
NFE & 1 & 2 & 3 & 4 & 5 \\
\midrule
Success rates & 49 & 48 & 51.5 & 79 & 84.5 \\
\midrule
NFE & 6 & 7 & 8 & 9 & 10 \\
\midrule
Success rates & 83.5 & 83.5 & 86 & 84.5 & 85 \\
\bottomrule
\end{tabular}
\end{table}

\subsubsection{Action Chunk Size}

To improve the model’s success rate under the one-step generation setting (NFE=1), we investigate the effect of the action chunk size on overall performance. The action chunk size $(n)$ defines the number of consecutive timesteps in each predicted action segment.

As shown in \cref{tab:chunksize}, both excessively small and excessively large chunk sizes lead to performance degradation.
We attribute this observation to the inherent balance between temporal consistency and observation freshness. When the chunk size is too large (\eg, n = 50), the model predicts an overly long sequence of future actions based on a single observation, this reduces responsiveness to environmental changes, causing accumulated errors and lower control precision. Conversely, when the chunk size is too small (\eg, n = 1), the model updates actions too frequently, resulting in only slight variations in the predicted actions between consecutive steps. This causes the robot to execute nearly redundant micro-actions without meaningful progress, yielding a success rate close to zero.

At $n=20$, the model achieves the highest performance under the one-step generation setting (NFE=1). This parameter strikes an optimal balance between temporal consistency and environmental responsiveness. The generated action sequences are temporally coherent and dynamically smooth, and the one-step generation process enables rapid inference while maintaining a high task success rate.

\begin{table}[htbp]
\centering
\caption{Success rates (\%) under various action chunk size with $flow\text{-}ratio$=0.2, Gamma=0.5 (NFE = 1).}
\label{tab:chunksize}
\begin{tabular}{@{}cc@{}}
\toprule
Chunk Size & Success Rates\\
\midrule
1 & 0 \\
10 & 82 \\
20 & 84.25 \\
50 & 50.5 \\
\bottomrule
\end{tabular}
\end{table}

\subsection{Ablation Study}

\begin{figure}[htbp]
  \centering
  \includegraphics[width=\columnwidth]{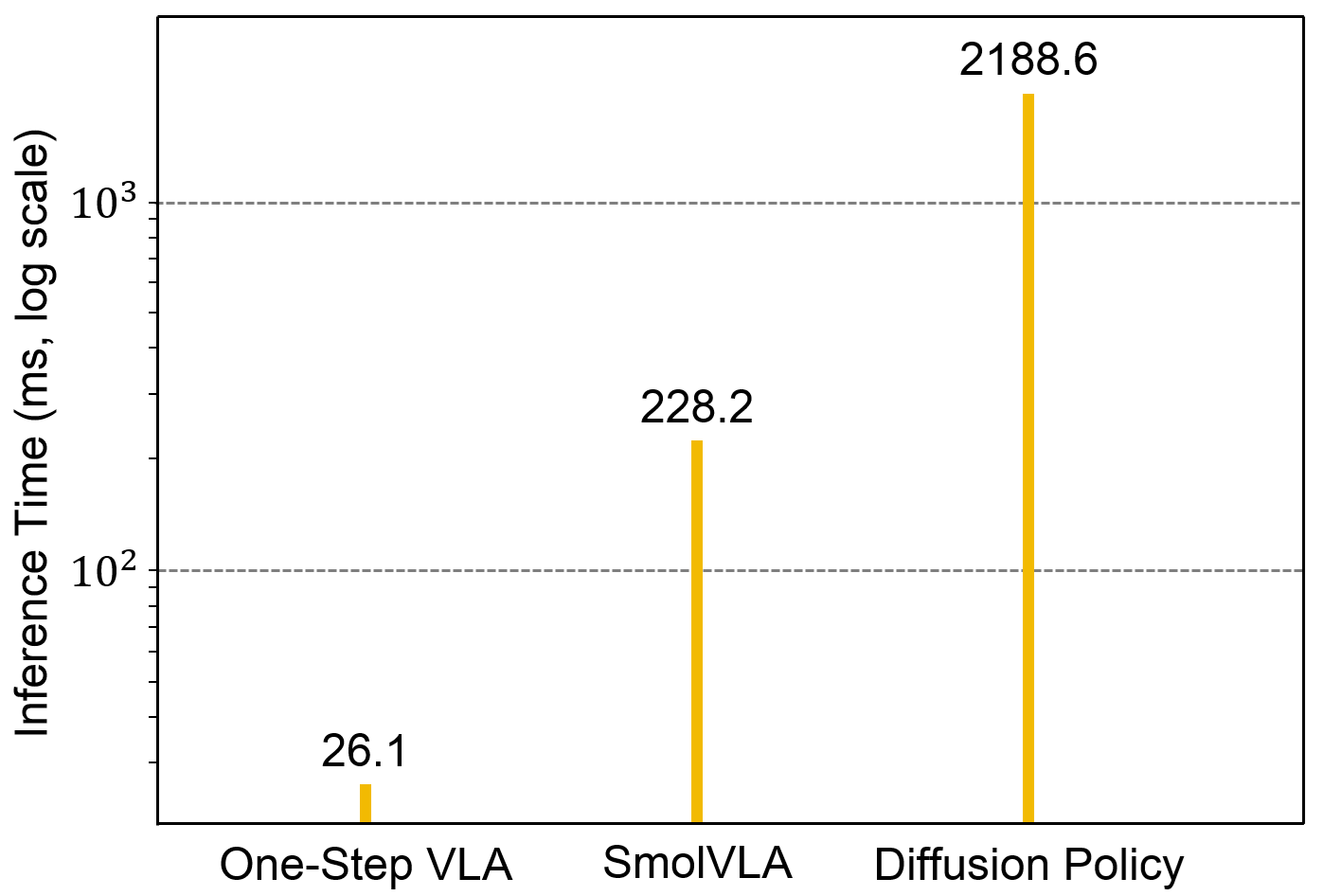}
  \caption{Illustrations showing the average action-generation speed of Diffusion Policy, SmolVLA, and One-Step VLA across three real manipulation tasks. The One-Step VLA is 8.7 times faster than SmolVLA and 83.9 times faster than Diffusion Policy.}
  \label{fig:inferencetime}
\end{figure}

One-Step VLA and SmolVLA are trained with mixed multi-task data, whereas Diffusion Policy is trained in a single-task setting. For evaluation, we conduct 10 independent rounds per task, each with 10 trials, and report the mean success rate and mean action-generation time.

Under the one-step configuration (NFE=1), the proposed Mean-Flow based One-Step VLA ($flow\text{-}ratio$=0.2, $\gamma$=0.5, chunk size=20) demonstrates a substantial efficiency advantage. As shown in \cref{fig:inferencetime} and  \cref{tab:abalation}, One-Step VLA is 8.7 and 83.9 times faster over SmolVLA and Diffusion Policy, respectively, while maintaining competitive task success rates.

This acceleration stems from fundamental differences in generation mechanisms. Diffusion Policy relies on iterative denoising steps, requiring multiple network evaluations per action chunk. FlowMatching-based VLA (\eg, SmolVLA) reduces computation cost but still rely on multi-step Euler integration, with each step involving a forward pass through the model. In contrast, One-Step VLA predicts the mean vector field over the entire time interval, enabling direct single-pass action generation without iterative refinement. In fact, eliminating repeated solver evaluations drastically reduces computational overhead, resulting in genuine one-step control.

We further observe that SmolVLA cannot reliably operate at NFE=1 without introducing additional consistency regularization, otherwise, trajectory amplitudes become unstable and unsuitable for deployment.

In the Stacking task, One-Step VLA achieves slightly lower success rates than SmolVLA, likely due to the high precision required for stacking, where multi-step refinement offers finer corrections. However, for PickPlace and long-horizon Sorting, One-Step VLA maintains comparable performance while delivering significantly faster inference, effectively narrowing the gap between one-step and multi-step action generation.

\begin{table}[htbp]
  \centering
  \caption{Success rates (\%) of three models evaluated across three different tasks.}
  \label{tab:abalation}
  \begin{tabular}{@{}lcccc@{}}
    \toprule
    Policy & PickPlace & Stacking & Sorting & Avg.\\
    \midrule
    \multicolumn{5}{@{}l}{\textbf{Single-task Training}} \\
    Diffusion Policy & 78.5 & 46 & 60 & 61.5\\
    \midrule
    \multicolumn{5}{@{}l}{\textbf{Multi-task Training}} \\
    SmolVLA          & 86.5 & 81.5 & 85.5 & 84.5\\
    One-Step VLA     & 88 & 64 & 82 & 78\\
    \bottomrule
  \end{tabular}
\end{table}

\section{CONCLUSIONS}

This paper proposes Mean-Flow based One-Step VLA, a new Vision–Language–Action framework that greatly increases generation speed and enables single-step action-sequence generation. The model predicts the mean vector field instead of the instantaneous one, allowing single-step inference that outputs actions after one evaluation. Real-robot experiments show that the proposed framework produces an action sequence with one forward pass, demonstrating inference speed of 8.7 and 83.9 times faster over SmolVLA and Diffusion Policy, respectively, while maintaining comparable success rates.

In future work, we will focus on improving the generalization and robustness of the Mean-Flow based One-Step VLA, especially for more complex tasks.

\addtolength{\textheight}{-12cm}   % This command serves to balance the column lengths

%%%%%%%%%%%%%%%%%%%%%%%%%%%%%%%%%%%%%%%%%%%%%%%%%%%%%%%%%%%%%%%%%%%%%%%%%%%%%%%%

%%%%%%%%%%%%%%%%%%%%%%%%%%%%%%%%%%%%%%%%%%%%%%%%%%%%%%%%%%%%%%%%%%%%%%%%%%%%%%%%

%%%%%%%%%%%%%%%%%%%%%%%%%%%%%%%%%%%%%%%%%%%%%%%%%%%%%%%%%%%%%%%%%%%%%%%%%%%%%%%%

%%%%%%%%%%%%%%%%%%%%%%%%%%%%%%%%%%%%%%%%%%%%%%%%%%%%%%%%%%%%%%%%%%%%%%%%%%%%%%%%

\bibliographystyle{ieeetr}
\bibliography{Mybib}
\end{document}